\documentclass{article}
\usepackage{kr}

\usepackage{times}
\usepackage{soul}
\usepackage{url}
\usepackage[hidelinks]{hyperref}
\usepackage[utf8]{inputenc}
\usepackage[small]{caption}
\usepackage{graphicx}
\usepackage{amsmath}
\usepackage{amsthm}
\usepackage{booktabs}
\usepackage{algorithm}
\usepackage{algorithmic}
\usepackage{amssymb}
\usepackage{color}
\usepackage{float}
\newfloat{mybox}{htbp}{lob}[section]
\floatname{mybox}{Box}

\newcommand{\nmi}{\,\mid\!\sim\,}

\title{Towards Dialogues for Joint Human-AI Reasoning and Value Alignment}

\author{%
Elfia Bezou-Vrakatseli\and
Oana Cocarascu\and
Sanjay Modgil
\affiliations
King's College London\\
\emails
\{elfia.bezou\_vrakatseli, oana.cocarascu, sanjay.modgil\}@kcl.ac.uk
}

\begin{document}

\maketitle

\begin{abstract}

We argue that enabling human-AI dialogue, purposed to support joint reasoning (i.e., ‘inquiry’), is important for ensuring that AI decision making is aligned with human values and preferences. In particular, we point to logic-based models of argumentation and dialogue, and suggest that the traditional focus on persuasion dialogues be replaced by a focus on inquiry dialogues, and the distinct challenges that joint inquiry raises. Given recent dramatic advances in the performance of large language models (LLMs), and the anticipated increase in their use for decision making, we provide a roadmap for research into inquiry dialogues for supporting joint human-LLM reasoning tasks that are ethically salient, and that thereby require that decisions are value aligned.
\end{abstract}

\section{Introduction}\label{Sec:Introduction}
As AI becomes more integrated into our lives, one needs to ensure both explainability of AI decisions, and that AI decisions
accommodate inputs from human users. This requirement is especially salient when decisions have ethical impact, and decisions need to be aligned with the preferences and values of human stakeholders. Indeed, addressing the so called `value alignment' problem \cite{Bostrom,russell2015research} -- ensuring that AI (in particular machine learning) systems achieve their goals in ways that are not anti-ethical to human preferences/values -- has acquired renewed urgency in light of recent dramatic advances in large language models (LLMs), and anticipated uses of LLMs in supporting decision making   \cite{wei2022emergent,abu2024supporting,papachristou2023leveraging,chen2023introspective,yang2023foundation}.

In this paper we make the following two claims so as to then advocate for a programme of research that: 1) advances work on argumentation-based dialogues; 2) enhances the reasoning and conversational capabilities of LLMs; 3) enables joint human-LLM reasoning.
Firstly, we claim that human-AI dialogue is required to support joint reasoning, enabling human input into decision making processes. In particular, these joint reasoning or `inquiry' dialogues should accommodate input and reasoning about human values and preferences. This will not only ensure that individual decisions are value-aligned, but that humans can be supported in refining their ethical positions, and the values/preferences informing decisions can then be learnt by AI systems so as to inform their future value-aligned decisions. Secondly, we claim that  dialogues generalising   argumentative formalisations of non-monotonic reasoning \cite{Dun95,IntroStuctured} present the most suitable candidates for enabling joint human-LLM inquiry.

In Section \ref{Sec:Mot} we further develop these claims by reference to motivating examples, and then enumerate a non-exhaustive list of research challenges if one is to realise the long term goal of joint human-LLM inquiry for value alignment. In Section \ref{Sec:Prog} we review current progress towards realising these research challenges, and key next steps in the research programme. We conclude in Section \ref{Sec:Conc}.

\section{Motivating Research into Joint Inquiry for Value Alignment}\label{Sec:Mot}

The last two decades have witnessed significant advances in machine learning, and accompanying concerns that, as AI systems become more autonomous and sophisticated on a trajectory to AGI and beyond \cite{Bostrom}, the overriding imperative of maximising reward functions may result in unforeseen decisions and actions that are harmful  to humans \cite{Bostrom,Alignment}. One of the most prominent emerging solutions to this problem of `value alignment' is Russell et.al's  Cooperative Inverse Reinforcement Learning (CIRL) framework    \cite{10.5555/3157382.3157535,HumanCompatible}. CIRL proposes that AI systems remain in a constant state of uncertainty about what the human value function is, and so are perpetually engaged in the primary goal of learning the value function through observation, query, and instruction. The challenge is to then ensure that the AI simultaneously acts so as to realise other goals for which it has been tasked\footnote{CIRL can be interpreted as advocating `enculturation of' AI, analogous to how young children acquire an understanding of values  through observation, questioning and instruction. Of course, the analogy breaks down when one considers that AI systems are  expected to perform complex tasks, often surpassing human capabilities, while simultaneously being `ethically enculturated'.}. As emphasised in \cite{modgil2017dialogical}, this entails that delegation of decision making tasks to AI systems requires simultaneously eliciting human input, and in particular ethically salient information that reveals stakeholder preferences and values; information that can then be learnt in order to inform future decisions. Consider the following simple example of an LLM specialised to serve as a personal assistant (PA) for a human user (H):

 \begin{quote}
\textbf{PA}: ``Given your budget and preferences for landscape and weather, I propose two options: (a) 7 days at the Dubai Beach Resort; (b) 7 days at the Miami Beach Resort.''

\textbf{H}: ``I prefer (b).''

\textbf{PA}: ``Why?''

\textbf{H}: ``Because according to Amnesty International, Dubai has unethical labour practices.''

\textbf{PA}: ``Ok. I will prepare an itinerary for Miami.''
\end{quote}

This short dialogue illustrates how, based on  \textbf{H}'s provision of factual information (i.e., budget constraints) and aesthetic preferences (weather and landscape), \textbf{PA} proposes two options and then elicits an ethically salient preference from \textbf{H}, effectively leading to a joint value aligned decision in favour of option (b), and ideally,  \textbf{PA} leveraging said ethical preference in later decisions.
Any such reasoning process should ideally be prescriptively governed by  rational principles for non-monotonic reasoning, wherein, as new information comes to light or there are conflicting reasons for alternative claims (inferences), preferences and values can be used to arbitrate in favour of a claim over its contradictory.

We thus turn to argumentative formalisations of non-monotonic (\textit{nm}) reasoning,  wherein arguments constructed (via chaining premises to a claim through application of deductive and/or default inference rules) from a given  individual's belief base $\mathcal{B}$, are related by a defeat relation. For example, if argument $A$'s claim contradicts argument $B$'s claim, then $A$ and $B$ attack each other. $A$  successfully attacks (i.e., `defeats') $B$, only if $A \nprec B$ (where $\prec$ is a preference relation over the constructed arguments). The argument framework ($AF$) of arguments related by defeats can then be evaluated under various Dung semantics \cite{Dun95}, so as to identify the justified (winning) arguments. The claims of these arguments can then be shown to equate with   various  \textit{nm} logical  consequence relations defined directly over $\mathcal{B}$ (e.g., see \cite{BDKT97,modgil2018abstract,ThangPS,young16}). Argument game proof theories (e.g., \cite{ModCam}) can then be deployed to establish whether a given argument in an $AF$ is justified under a given semantics, and hence whether its claim is an \textit{nm} consequence from the underlying $\mathcal{B}$. These argument game proof theories can be   adapted \cite{modgil2017dialogical} to yield \textit{dialogical formalisations} of  \textit{nm} reasoning (e.g., \cite{Fan201420,prakken2005coherence}) in which  agents can exchange locutions -- not just arguments but claims, questions etc. -- that conform to protocols generalising  rules in the aforementioned argument games. The protocols and evaluation of the resultant \emph{graph of locutions} are defined so that under certain conditions:
\begin{quote}
a communicated claim $\alpha$ is established by dialogue $d$ iff $\alpha$ is the  claim of a justified argument in $AF_{\mathcal{B}}$
\end{quote}

\noindent where instead of $AF_{\mathcal{B}}$ being defined by a static given belief base $\mathcal{B}$, as in the paradigmatic case of single agent reasoning,
$AF_{\mathcal{B}}$ is   \textit{incrementally}  defined either by the publicly asserted beliefs $\mathcal{B}$ asserted in the dialogue $d$, or the union of the belief bases of the agents participating in $d$. One thus obtains (given the aforementioned soundness and completeness results for argumentative characterisations of \textit{nm} consequence relations)  communicative accounts of distributed, i.e., `joint', \emph{nm} reasoning ($\nmi$ below denotes the consequence relation of some \textit{nm} logic):\begin{quote} a communicated claim $\alpha$ is established by dialogue $d$ iff $\mathcal{B} \nmi \alpha$   \hspace{8mm}
\end{quote}

Consider our example dialogue. The budget and aesthetic information provided by \textbf{H}, combined with other travel related information, is used as premises in two mutually attacking arguments $A$ and $B$, respectively claiming the Dubai and Miami options.  \textbf{PA} claims a preference for $B$, and upon questioning, argues against (i.e., defeats) $B$ with an argument $C$ claiming unethical labour practices in Dubai. Hence $A$ is defended by $C$'s defeat on $B$ and the   dialogue establishes option (b).

Key to enabling such dialogues is the use of argument schemes (\textit{AS})
and critical questions (\textit{CQs}) \cite{walt} that have been developed within the field of informal logic.   \textit{AS} are templates capturing stereotypical patterns of intra-argument reasoning, and can be instantiated    by natural and/or formal logic languages. For example, one can suppose  both $A$ and $B$  to be instances of the \textit{Argument for Action (AA)} scheme \cite{ATKINSON2007855}: `In circumstances $S$, doing action $Act$ will achieve goal $G$ and so promote value $V$' (where $S$ includes, amongst other things, budget  and aesthetic information, $G$ is the goal of `going on holiday', $V$ the value of `pleasure' and $Act$ the respective decision options). Associated with \textit{AA} are sixteen critical questions which identify possible criteria for attacking, or indeed supporting, any concrete instance of \textit{AA}. For example, the \textit{CQ} `Is it the case that $S$ ?' can be addressed as a question by agent $Ag$ to an agent $Ag'$  who has submitted an argument $X$ instantiating  \textit{AA}. In response, $Ag'$ is dialectically obligated to provide an argument
claiming $S$.\footnote{E.g., an argument claiming $Act$ = `lockdown' may cite $S$ = `there is a pandemic', and is not justified until such point that in response to questioning whether it is the case that there is a pandemic, relevant evidence is cited to support this claim.} \textit{CQs} can also be used to challenge arguments. In our running example, \textbf{H} uses the
\textit{CQ} `Does action $A$ demote some other value' to submit the argument claiming unethical labour practices.
$C$ is itself an instance of  the argument from expert opinion (\textit{EO}) scheme which cites an expert having asserted a claim as support for the claim. \textit{EO} has its own \textit{CQ} which can similarly be used to respond to $C$. Thus,
\textit{AS} and \textit{CQ}s help map out the space of possible reasoning, and a key requirement for realising the use of \textit{AS} and \textit{CQs} in dialogues is the following:

\begin{quote} \textit{Strategic considerations should inform the choice of \textit{AS} or \textit{CQs} at any given point in a dialogue}  (\textbf{$R_{strat}$})
\end{quote}

Suppose that instead of $C$, \textbf{H} submitted the locution $C'$:  ``Dubai has unethical labour practices''.
Now note that $C'$ is an incomplete argument -- an `\textit{enthymeme}'  \cite{walton2009argumentation} -- consisting only of a claim. Use of enthymemes are characteristic of real world dialogues, and ideally, the dialogue would evolve so that either the \textbf{PA} or \textbf{H} is prompted to supply evidence in support of the claim, to, as it were, `flesh out' the claim $C'$ into a fully fledged argument $C$. Thus, \textbf{PA} might question `why $C'$', so that \textbf{H} is prompted to submit the expert opinion as a premise. Indeed,  \textbf{H} may herself submit $C'$ as a query;  she believes $C'$ to be the case, but is inviting \textbf{PA} (or perhaps some other agent) to provide the missing premise and so complete $C'$ into an argument. In either case, until the dialectical obligation is met, $C'$ does not constitute a valid defeat on $A$,  and $B$'s claim is not established. Hence the following requirement for realising inquiry dialogues:

\begin{quote} \textit{Dialogues should be guided so that declarative locutions can be combined to define complete arguments} (\textbf{$R_{enth}$})
\end{quote}

Consider the following example dialogue\footnote{Admittedly, this does not represent a typical decision making scenario; rather it is used to illustrate requirements for ethical reasoning.}  between a human \textbf{H} and a `Socratic LLM' \textbf{SocrAPP}:

            \begin{quote}

            1. \textbf{H}: ``I propose that we should make abortion illegal.''

            2. \textbf{SocrAPP}: ``Why?''

            3. \textbf{H}: ``Because it amounts to the early termination of life.''

            4. \textbf{SocrAPP}: ``So?''

            5. \textbf{H}: ``All humans have a right to life.''

            6. \textbf{SocrAPP}: \raggedright ``What about the right of a woman to choose?''

            7. \textbf{H}: ``What do you mean?''

            8. \textbf{SocrAPP}: ``Your argument appeals to the right to life, but making abortion illegal violates the right of a woman to choose.''

            9. \textbf{H}: ``But the right to life takes precedence over the right to choose.''

            10. \textbf{SocrAPP}: ``Why?''

             :

            n ($>$ 11). \textbf{SocrAPP}: ``Here are links to studies showing that in developing countries, giving women the right to choose reduces poverty . Here are some articles that discuss and analyse these findings.''\\
            \end{quote}

In 1--5, SocrAPP elicits (in 2 and 4) information from H. This information  completes the  initial claim (the enthymeme in 1) into an  argument $A$, instantiating a variant of scheme \textit{AA} -- which we call \textit{AA(R$^+$)} -- that appeals to a right (to life) rather than a value. In 6, SocrAPP uses a \textit{CQ} for  \textit{AA(R$^+$)} -- \textit{Does the action violate some other right?} -- to
submit an
enthymeme for an argument $B$ instantiating a scheme (\textit{AA(R$^-$)}) prohibiting   an action if it violates an established right. We note that schemes such as \textit{AA(R$^+$)} and \textit{AA(R$^-$)} and the  \textit{CQ} we posit for the former, are novel, and specialised for ethical reasoning. Indeed, experience with the use of \textit{AS} $\&$  \textit{CQs} for guiding real-world deliberation in the medical domain \cite{tolchinsky2012deliberation} highlights the limiting generality of \textit{AS} $\&$  \textit{CQs}  \cite{walt}, and the need for \textit{AS} $\&$  \textit{CQs} specialised for medical reasoning. Hence we posit the following requirement for joint inquiry dialogues that facilitate value alignment:

\begin{quote} \textit{Schemes and Critical Questions specialised for ethical reasoning} (\textbf{$R_{ScCQ_{Ethical }}$})
\end{quote}

In response to H's query in 7, SocrAPP fulfills an explanatory remit (in 8), completing 6 into the argument $B$, and additionally making explicit that $B$ and $A$ attack each other. In line 9, H declares a preference for the right to life over the right to choose (i.e., the preference $B \prec A$) so that only $A$ defeats $B$ (the dialogue has therefore thus far established H's initial claim). In line 10, SocrAPP queries H's rationale for this preference. H's initial claim is now `dis-established', and  H is now required to provide a `metalevel' argument for why one argument is preferred to another in order to re-stablish her initial claim.

Now, recall that we have thus far proposed \textit{sound and complete} inquiry dialogues, in the sense that evaluation of a graph of locutions establishes a claim $\alpha$ iff
$\alpha$ is the claim of a justified argument in the $AF$ constructed from $\mathcal{B}$ (the   contents of  declarative locutions exchanged, or of the participating agents' belief bases), which in turn equates with
$\mathcal{B} \nmi  \alpha$ in some \textit{nm} logic. However, in real-world (especially ethically salient) dialogues, one  does not always expect consensus as to the preferences used to determine the success of attacks and defeats; rather one expects reasoning about possibly contradictory preferences, and argumentation based resolution of these conflicts. Hence, one requires:

\begin{quote} \textit{Metalevel}
Schemes and Critical Questions specialised for metalevel reasoning (\textbf{$R_{ScCQ_{Meta}}$})\footnote{  \cite{kokciyan2021applying} also posit  \textit{meta-schemes}  tailored to reasoning about other object-level features of $AF$, including e.g., the rationale for why arguments attack each other, when said rationale is not based on negation, e.g., when arguments for two actions conflict because they have mutually exclusive pre-conditions.}
\end{quote}

and

\begin{quote} \textit{Soundness and Completeness of dialogues that accommodate argumentation based metalevel reasoning about preferences and values} (\textbf{$R_{SC-Pref}$})
\end{quote}

Let us suppose the dialogue continues,  but (as is typically the case when appealing to rights), there is no way to resolve the conflict between arguments expressing contradictory preferences over rights. The claim is then subject to consequentialist evaluation, and in   line $n$, SocrAPP is responding to a prompt requesting evidence for the positive and negative consequences of legalising/prohibiting abortion (said evidence instantiating further variants of the $AA$ scheme that justify actions on consequentialist grounds).

An important feature of our proposed inquiry dialogues for value alignment is that they facilitate rational exploration of the space of possible ethical evaluations one might take with respect to a decision. As illustrated above, the implication is that although a user might be committed  to a rights-based ethical perspective, they would ideally be exposed to alternative viewpoints, such as consequentialist criteria, when evaluating decisions. Indeed, recent philosophical approaches eschew a `one size fits all' approach, and instead advocate judgements that aggregate evaluations of decision options under different ethical theories
\cite{macaskill2020moral,szabo2024moral}. Hence, the requirement for object-level domain specific ethical \textit{AS} and \textit{CQs}, and metalevel \textit{AS} and \textit{CQs} that may be deployed to aggregate the evaluations yielded by object-level arguments. Indeed, the implication is that a user may be encouraged to refine or even revise their ethical preferences, with the proviso that the AI interlocutor's role is not to persuade the user to commit to a particular ethical judgement, but rather to explore the space of ethical reasoning. This then implies use of other dialectical moves, such as the well known Socratic move demonstrating that an interlocutor has contradicted herself \cite{vlastos1982socratic}. For example, suppose that after line 5, SocrAPP queries whether H is for or against the death  penalty, and H replies to say she is in favour. SocrAPP can then reveal that H has contradicted herself (given her appeal to the right to life), prompting H to refine her ethical stance: ``All humans, \textit{except those that take the life of another}, have the right to life''. Hence the requirement that:

\begin{quote} \textit{Dialogues accommodate dialectical moves other than the moving of attacking/supporting arguments, and results in changes to the beliefs of interlocutors} (\textbf{$R_{Dia}$})
\end{quote}

In this section we have, through the use of illustrative examples, enumerated a non-exhaustive list of requirements for  inquiry dialogues that formalise joint non-monotonic reasoning; dialogues purposed to accommodate human ethical preferences and values that may be refined and subsequently learnt by AI interlocutors that are simultaneously tasked with supporting human users in their decision making. In the next section we briefly review progress towards realising these requirements, and then focus on challenges that need to be addressed if LLMs are to successfully engage in such dialogues.

\section{Progress Towards Realising Joint Inquiry for Value Alignment}\label{Sec:Prog}

\subsection{Towards Argumentation-based Dialogues for Value Aligned Inquiry}

There is a sparsity of work on formalising
Inquiry Dialogues 
 such that the aforementioned soundness and completeness results hold. A notable exception is the work of \cite{BlackAutonomousAA} in which agents can challenge arguments and collaboratively construct arguments in the language of Defeasible Logic Programming \cite{SIMARI_2004}. Rather, the majority of research focuses on argumentation-based `zero sum' \textit{persuasion} dialogues \cite{Prakken2009} that are inherently adversarial, and thus clearly unsuitable given the collaborative goal of joint reasoning that we are interested in. There are then key requirements specific to inquiry that do not arise in the case of persuasion. Notably, arguments are constructed by individual agents in the latter, whereas inquiry deploys the epistemic resources of multiple agents in constructing individual arguments. Moreover, in persuasion dialogues, an agent cannot target their own argument with defeats, whereas inquiry should licence targeting one's own argument, given that the aim is to get to `truth of the matter' in respect of factual information, and to support decision options that are rationally justified as  being optimal for users. Indeed, given these collaborative aims, the assumptions under which one expects the previous section's soundness and completeness results for dialogues to hold -- namely that interlocutors are honest and are exhaustive in submitting locutions licensed by the protocol -- are reasonable.
In light of what we perceive to be a dearth of formal models for sound and complete inquiry dialogues, we briefly point to progress made towards realising the requirements enumerated in Section \ref{Sec:Mot}.

Firstly, the \textit{ASPIC}$^+$ framework \cite{modgil2013general}  for formalising argumentative characterisation of non-monotonic reasoning, has been extended to accommodate metalevel reasoning about preferences and values \cite{modgil2013general,EAFandASPIC}. Dialogical formulations of this extended framework have been developed
\cite{modgil2017towards}, but it remains to accommodate moves distinctive of inquiry rather than persuasion (e.g. allowing multiple interlocutors to collaboratively construct individual arguments), and to show soundness and completeness results (\textbf{$R_{SC-Pref}$}). Moreover, enthymemes are not accommodated, but sound and complete inquiry dialogues that accommodate enthymemes and are based on the \textit{standard} \textit{ASPIC}$^+$ framework have been developed \cite{xydis2020enthymemes,xydis2022sound} (addressing \textbf{$R_{enth}$}), and integration of this work with \cite{modgil2017towards}  represents a significant next research objective. Finally, recent argumentative formalisations of non-monotonic reasoning have been developed to accommodate attacks that dialectically demonstrate the inconsistency of premises of challenged arguments \cite{d2018classical,RationalASPIC}, partly addressing \textbf{$R_{Dia}$} (it remains to formalise changes to beliefs in response to such moves).

Advances have also been made in addressing \textbf{$R_{ScCQ_{Ethical }}$}, as detailed in \cite{bezou2022towards}, in which  prevalent \textit{AS} used in ethical reasoning on the debate platform Kialo (\url{https://www.kialo.com}) are identified and have been used to guide  formulation of a new, specialized taxonomy for ethical reasoning. The (ongoing) creation of this taxonomy categorises common \textit{AS} used in ethical debates and introduces new \textit{AS} specifically tailored for such discussions. Furthermore, this taxonomy proposes meta-schemes (\textbf{$R_{ScCQ_{Meta}}$}) that facilitate reasoning about and comparing ethical evaluations yielded by object level arguments instantiating ethical schemes.



 \subsection{Towards Large Language Model Interlocutors}

Large Language Models (LLMs)
~\cite{brown2020language,hoffmann2022training,touvron2023llama,zhang2022opt} have significantly advanced the state-of-the-art on a wide range of language understanding tasks, by harnessing vast amounts of data,
and excel in generating coherent and relevant text based on minimal prompts.
This technological leap has ushered in a new era of AI, where machines can engage in more human-like conversations and provide insights with unprecedented accuracy \cite{crispino2023agent,hamalainen2023evaluating,castagna2023eqrbot,lu2023memochat,bubeck2023sparks,liu2024llm,lee-etal-2024-towards,10.1145/3616855.3635856,castagna2024computational}, with multiple studies investigating the limits of their abilities, including
the ability to learn \cite{brown2020language}, reason \cite{bubeck2023sparks,wei2022chain,wang2022self,kiciman2023causal}, and comprehend intricate language structures such as metaphors \cite{wachowiak2023does}.
The remarkable capabilities of LLMs have led to their widespread adoption across various fields, including finance \cite{araci2019finbert,wu2023bloomberggpt}, robotics \cite{driess2023palm,huang2023instruct2act}, medicine \cite{huang2019clinicalbert,bolton2022biomedlm}, and even crisis counselling \cite{lee-etal-2024-towards}.

However, despite their success,
LLMs face challenges when faced with complex reasoning tasks, especially involving knowledge and symbolic reasoning \cite{zhao2023survey}, and exhibit deficiencies in both planning and reasoning abilities \cite{bubeck2023sparks}. Dialogue, specifically, remains a challenging task for LLMs \cite{zhao2023survey,wang2023survey,sumers2023cognitive,xi2023rise,liu2024llm}.
In order for LLMs to engage in joint inquiry as described in the previous sections, LLMs must be able to instantiate \textit{AS} and \textit{CQs} effectively. We discuss below a non-exhaustive selection of challenges for realising this goal.


Firstly, LLMs need to be able to classify arguments using \textit{AS} and to identify how one argument relates to another, through the use of their \textit{CQs}. This is an open challenge but, given their impressive capabilities, there is now the potential to overcome previous hurdles in AI natural language processing, particularly regarding the instantiation of \textit{AS} and \textit{CQs}. Indeed, our preliminary investigations suggest that not only are  LLMs capable of classifying arguments according to the \textit{AS}   they instantiate, but are also able to articulate why said \textit{AS} have been identified. However, as with human annotators of argumentative corpus \cite{waltmac,habernal2017argumentation,viss}, LLMs struggle to differentiate between similar \textit{AS}.

Building on these investigations suggests a number of  open challenges. Firstly, harnessing the abilities of LLMs to identify arguments that support or challenge a given claim, the schemes these arguments instantiate, and hence the ability to flesh out enthymemes (recall   \textbf{$R_{enth}$}). Secondly, while LLMs can be promted to generate example dialogues on a given topic, the ability to identify the critical questions that arguments are responding to is still limited.

Moreover, effective instantiation of \textit{AS} and \textit{CQs} requires LLMs to not only recognise,
but also apply these reasoning structures, in a \textit{contextually
appropriate} manner. The challenges discussed above go hand in hand with the LLMs’ need to retain conversational histories over prolonged interactions.
Maintaining conversational histories in a short-term memory module is currently an active research area \cite{lu2023memochat,zhang2024survey},
which could be leveraged to
maintain coherence across extended dialogues and multiple interactions.

Finally, one would want LLMs to develop reasoning capacities enabling more complex dialectical moves (\textbf{$R_{Dia}$}), and to strategise (\textbf{$R_{strat}$}). Besides identifying points of support and attack, there is a need for subsequent instantiation of \textit{AS} to construct arguments to support/attack, supported by a \textit{dialogue manager}, with evaluation supported by a \textit{protocol manager}. To effectively mimic human-like discourse management, and address \textbf{$R_{strat}$}, LLMs should additionally be able to navigate the flow of conversation, deciding when to probe deeper or when to guide towards conclusions. This can be viewed as a search problem in a game tree, where the choice of moves amounts to the choice of \textit{AS} and \textit{CQs}.


One technique that has been developed to improve the performance and maximize the utility of LLMs is \textit{prompt engineering} (e.g., \cite{crispino2023agent,liu2023pre,brown2020language}), which is increasingly essential for proficient interaction with LLMs \cite{white2023prompt,liu2022design}.
Prompts are instructions given to an
LLM by customizing it and/or enhancing its capabilities to enforce rules, automate processes, and ensure desired attributes and volumes of generated content \cite{crispino2023agent,liu2023pre}.
Effectively, prompt engineering is the means by which LLMs are `programmed' via prompts, dictating how the models should process information.

Chain of thought (CoT) prompting  \cite{wei2022chain}, which consists in breaking down an initial task/question into intermediate reasoning steps, holds promise for addressing the challenges we have identified. Overall, intermediate reasoning has been shown to improve LLMs' performance \cite{rajani2019explain} and CoT is currently used for improving the reasoning abilities of LLMs \cite{turpin2024language,zelikman2022star,lewkowycz2022solving,suzgun2022challenging}.
Essentially, CoT represents a series of intermediate natural language reasoning steps; for instance, \textit{step-by-step} \cite{kojima2022large}, which guides the model through a series of reasoning steps, or \textit{decomposition} \cite{zhang2022automatic}, where complex problems are broken down into simpler sub-problems.
Especially promising for joint inquiry is \textit{maieutic prompting} \cite{jung2022maieutic,10099179} which, building on the two aforementioned types, utilises a recursive dialogue-based prompting, emphasising logically consistent reasoning through recursive explanations at each step. This may be\ helpful for addressing \textbf{$R_{Dia}$} and supporting LLMs in developing the reasoning capacities for addressing different dialectical moves.






\section{Conclusions}\label{Sec:Conc}

This paper advocates for a focus on argumentation-based inquiry dialogues so as to integrate human ethical preferences in joint human-LLM reasoning and decision making. We have argued that such dialogues are warranted by state-of-the-art proposals for value alignment, and moreover support humans in refining their ethical positions. We have enumerated a number of research challenges for realising said dialogues, in the areas of computational argument and dialogue, and LLMs, and pointed to progress made.
 While recent advancements in LLMs offer promising capabilities, significant challenges in dialogue management persist. Addressing these challenges will be key to developing AI systems that can engage effectively in joint inquiry for value alignment.

\section*{Acknowledgments}
 This work was supported by the UK Research and Innovation Centre for Doctoral Training in Safe and Trusted Artificial Intelligence [grant number EP/S023356/1].

\bibliographystyle{kr}
\bibliography{KR24}

\begin{thebibliography}{}

\bibitem[\protect\citeauthoryear{Abbasiantaeb \bgroup et al\mbox.\egroup
  }{2024}]{10.1145/3616855.3635856}
Abbasiantaeb, Z.; Yuan, Y.; Kanoulas, E.; and Aliannejadi, M.
\newblock 2024.
\newblock Let the llms talk: Simulating human-to-human conversational qa via
  zero-shot llm-to-llm interactions.
\newblock In {\em Proceedings of the 17th ACM International Conference on Web
  Search and Data Mining}, WSDM '24,  8–17.
\newblock New York, NY, USA: Association for Computing Machinery.

\bibitem[\protect\citeauthoryear{Abu-Rasheed \bgroup et al\mbox.\egroup
  }{2024}]{abu2024supporting}
Abu-Rasheed, H.; Abdulsalam, M.~H.; Weber, C.; and Fathi, M.
\newblock 2024.
\newblock Supporting student decisions on learning recommendations: An
  llm-based chatbot with knowledge graph contextualization for conversational
  explainability and mentoring.
\newblock {\em arXiv preprint arXiv:2401.08517}.

\bibitem[\protect\citeauthoryear{Araci}{2019}]{araci2019finbert}
Araci, D.
\newblock 2019.
\newblock Finbert: Financial sentiment analysis with pre-trained language
  models.
\newblock {\em arXiv preprint arXiv:1908.10063}.

\bibitem[\protect\citeauthoryear{Atkinson and
  Bench-Capon}{2007}]{ATKINSON2007855}
Atkinson, K., and Bench-Capon, T.
\newblock 2007.
\newblock Practical reasoning as presumptive argumentation using action based
  alternating transition systems.
\newblock {\em Artificial Intelligence} 171(10):855--874.

\bibitem[\protect\citeauthoryear{Besnard \bgroup et al\mbox.\egroup
  }{2014}]{IntroStuctured}
Besnard, P.; Garcia, A.; Hunter, A.; Modgil, S.; Prakken, H.; Simari, G.; and
  Toni, F.
\newblock 2014.
\newblock Introduction to structured argumentation.
\newblock {\em Argument and Computation} 5.

\bibitem[\protect\citeauthoryear{Bezou-Vrakatseli, Cocarascu, and
  Modgil}{2022}]{bezou2022towards}
Bezou-Vrakatseli, E.; Cocarascu, O.; and Modgil, S.
\newblock 2022.
\newblock Towards an argument scheme classification for ethical reasoning.
\newblock In Grasso, F.; Green, N.~L.; Schneider, J.; and Wells, S., eds., {\em
  Proceedings of the 22nd Workshop on Computational Models of Natural Argument,
  CMNA@COMMA 2022, Cardiff, Wales, September 12, 2022}, volume 3205 of {\em
  {CEUR} Workshop Proceedings},  13--17.
\newblock CEUR-WS.org.

\bibitem[\protect\citeauthoryear{Black and Hunter}{2009}]{BlackAutonomousAA}
Black, E., and Hunter, A.
\newblock 2009.
\newblock An inquiry dialogue system.
\newblock {\em Autonomous Agents and Multi-agent Systems} 19:173–209.

\bibitem[\protect\citeauthoryear{Bolton \bgroup et al\mbox.\egroup
  }{2022}]{bolton2022biomedlm}
Bolton, E.; Hall, D.; Yasunaga, M.; Lee, T.; Manning, C.; and Liang, P.
\newblock 2022.
\newblock Biomedlm: a domain-specific large language model for biomedical text.
\newblock {\em Stanford CRFM Blog}.

\bibitem[\protect\citeauthoryear{Bondarenko \bgroup et al\mbox.\egroup
  }{1997}]{BDKT97}
Bondarenko, A.; Dung, P.; Kowalski, R.; and Toni, F.
\newblock 1997.
\newblock An abstract, argumentation--theoretic approach to default reasoning.
\newblock {\em Artificial Intelligence} 93:63--101.

\bibitem[\protect\citeauthoryear{Bostrom}{2014}]{Bostrom}
Bostrom, N.
\newblock 2014.
\newblock {\em Superintelligence: Paths, Dangers, Strategies}.
\newblock Oxford University Press.

\bibitem[\protect\citeauthoryear{Brown \bgroup et al\mbox.\egroup
  }{2020}]{brown2020language}
Brown, T.; Mann, B.; Ryder, N.; Subbiah, M.; Kaplan, J.~D.; Dhariwal, P.;
  Neelakantan, A.; Shyam, P.; Sastry, G.; Askell, A.; et~al.
\newblock 2020.
\newblock Language models are few-shot learners.
\newblock {\em Advances in neural information processing systems}
  33:1877--1901.

\bibitem[\protect\citeauthoryear{Bubeck \bgroup et al\mbox.\egroup
  }{2023}]{bubeck2023sparks}
Bubeck, S.; Chandrasekaran, V.; Eldan, R.; Gehrke, J.; Horvitz, E.; Kamar, E.;
  Lee, P.; Lee, Y.~T.; Li, Y.; Lundberg, S.; et~al.
\newblock 2023.
\newblock Sparks of artificial general intelligence: Early experiments with
  gpt-4.
\newblock {\em arXiv preprint arXiv:2303.12712}.

\bibitem[\protect\citeauthoryear{Castagna \bgroup et al\mbox.\egroup
  }{2023}]{castagna2023eqrbot}
Castagna, F.; Garton, A.; McBurney, P.; Parsons, S.; Sassoon, I.; and Sklar,
  E.~I.
\newblock 2023.
\newblock Eqrbot: A chatbot delivering eqr argument-based explanations.
\newblock {\em Frontiers in Artificial Intelligence} 6:1045614.

\bibitem[\protect\citeauthoryear{Castagna \bgroup et al\mbox.\egroup
  }{2024}]{castagna2024computational}
Castagna, F.; Kokciyan, N.; Sassoon, I.; Parsons, S.; and Sklar, E.
\newblock 2024.
\newblock Computational argumentation-based chatbots: a survey.
\newblock {\em arXiv preprint arXiv:2401.03454}.

\bibitem[\protect\citeauthoryear{Chang}{2023}]{10099179}
Chang, E.~Y.
\newblock 2023.
\newblock Prompting large language models with the socratic method.
\newblock In {\em 2023 IEEE 13th Annual Computing and Communication Workshop
  and Conference (CCWC)},  0351--0360.

\bibitem[\protect\citeauthoryear{Chen \bgroup et al\mbox.\egroup
  }{2023}]{chen2023introspective}
Chen, L.; Wang, L.; Dong, H.; Du, Y.; Yan, J.; Yang, F.; Li, S.; Zhao, P.; Qin,
  S.; Rajmohan, S.; et~al.
\newblock 2023.
\newblock Introspective tips: Large language model for in-context decision
  making.
\newblock {\em arXiv preprint arXiv:2305.11598}.

\bibitem[\protect\citeauthoryear{Crispino \bgroup et al\mbox.\egroup
  }{2023}]{crispino2023agent}
Crispino, N.; Montgomery, K.; Zeng, F.; Song, D.; and Wang, C.
\newblock 2023.
\newblock Agent instructs large language models to be general zero-shot
  reasoners.
\newblock {\em arXiv preprint arXiv:2310.03710}.

\bibitem[\protect\citeauthoryear{D'Agostino and Modgil}{2018}]{d2018classical}
D'Agostino, M., and Modgil, S.
\newblock 2018.
\newblock Classical logic, argument and dialectic.
\newblock {\em Artificial Intelligence} 262:15--51.

\bibitem[\protect\citeauthoryear{D'Agostino and Modgil}{2021}]{RationalASPIC}
D'Agostino, M., and Modgil, S.
\newblock 2021.
\newblock A fully rational account of structured argumentation under resource
  bounds.
\newblock In {\em Proceedings of the Twenty-Ninth International Joint
  Conference on Artificial Intelligence}, IJCAI'20.

\bibitem[\protect\citeauthoryear{Driess \bgroup et al\mbox.\egroup
  }{2023}]{driess2023palm}
Driess, D.; Xia, F.; Sajjadi, M.~S.; Lynch, C.; Chowdhery, A.; Ichter, B.;
  Wahid, A.; Tompson, J.; Vuong, Q.; Yu, T.; et~al.
\newblock 2023.
\newblock Palm-e: An embodied multimodal language model.
\newblock {\em arXiv preprint arXiv:2303.03378}.

\bibitem[\protect\citeauthoryear{Dung}{1995}]{Dun95}
Dung, P.
\newblock 1995.
\newblock On the acceptability of arguments and its fundamental role in
  nonmonotonic reasoning, logic programming and $n$--person games.
\newblock {\em Artificial Intelligence} 77:321--357.

\bibitem[\protect\citeauthoryear{Fan and Toni}{2014}]{Fan201420}
Fan, X., and Toni, F.
\newblock 2014.
\newblock A general framework for sound assumption--based argumentation
  dialogues.
\newblock {\em Artificial. Intelligence.} 216(0):20 -- 54.

\bibitem[\protect\citeauthoryear{Garcia and Simari}{2004}]{SIMARI_2004}
Garcia, A.~J., and Simari, G.~R.
\newblock 2004.
\newblock Defeasible logic programming: an argumentative approach.
\newblock {\em Theory and Practice of Logic Programming} 4(1–2):95–138.

\bibitem[\protect\citeauthoryear{Habernal and
  Gurevych}{2017}]{habernal2017argumentation}
Habernal, I., and Gurevych, I.
\newblock 2017.
\newblock Argumentation mining in user-generated web discourse.
\newblock {\em Computational linguistics} 43(1):125--179.

\bibitem[\protect\citeauthoryear{Hadfield-Menell \bgroup et al\mbox.\egroup
  }{2016}]{10.5555/3157382.3157535}
Hadfield-Menell, D.; Dragan, A.; Abbeel, P.; and Russell, S.
\newblock 2016.
\newblock Cooperative inverse reinforcement learning.
\newblock In {\em Proceedings of the 30th International Conference on Neural
  Information Processing Systems}, NIPS'16,  3916–3924.
\newblock Red Hook, NY, USA: Curran Associates Inc.

\bibitem[\protect\citeauthoryear{H{\"a}m{\"a}l{\"a}inen, Tavast, and
  Kunnari}{2023}]{hamalainen2023evaluating}
H{\"a}m{\"a}l{\"a}inen, P.; Tavast, M.; and Kunnari, A.
\newblock 2023.
\newblock Evaluating large language models in generating synthetic hci research
  data: a case study.
\newblock In {\em Proceedings of the 2023 CHI Conference on Human Factors in
  Computing Systems},  1--19.

\bibitem[\protect\citeauthoryear{Hoffmann \bgroup et al\mbox.\egroup
  }{2022}]{hoffmann2022training}
Hoffmann, J.; Borgeaud, S.; Mensch, A.; Buchatskaya, E.; Cai, T.; Rutherford,
  E.; Casas, D. d.~L.; Hendricks, L.~A.; Welbl, J.; Clark, A.; et~al.
\newblock 2022.
\newblock Training compute-optimal large language models.
\newblock {\em arXiv preprint arXiv:2203.15556}.

\bibitem[\protect\citeauthoryear{Huang, Altosaar, and
  Ranganath}{2019}]{huang2019clinicalbert}
Huang, K.; Altosaar, J.; and Ranganath, R.
\newblock 2019.
\newblock Clinicalbert: Modeling clinical notes and predicting hospital
  readmission.
\newblock {\em arXiv preprint arXiv:1904.05342}.

\bibitem[\protect\citeauthoryear{Huang \bgroup et al\mbox.\egroup
  }{2023}]{huang2023instruct2act}
Huang, S.; Jiang, Z.; Dong, H.; Qiao, Y.; Gao, P.; and Li, H.
\newblock 2023.
\newblock Instruct2act: Mapping multi-modality instructions to robotic actions
  with large language model.
\newblock {\em arXiv preprint arXiv:2305.11176}.

\bibitem[\protect\citeauthoryear{Jung \bgroup et al\mbox.\egroup
  }{2022}]{jung2022maieutic}
Jung, J.; Qin, L.; Welleck, S.; Brahman, F.; Bhagavatula, C.; Bras, R.~L.; and
  Choi, Y.
\newblock 2022.
\newblock Maieutic prompting: Logically consistent reasoning with recursive
  explanations.
\newblock {\em arXiv preprint arXiv:2205.11822}.

\bibitem[\protect\citeauthoryear{K{\i}c{\i}man \bgroup et al\mbox.\egroup
  }{2023}]{kiciman2023causal}
K{\i}c{\i}man, E.; Ness, R.; Sharma, A.; and Tan, C.
\newblock 2023.
\newblock Causal reasoning and large language models: Opening a new frontier
  for causality.
\newblock {\em arXiv preprint arXiv:2305.00050}.

\bibitem[\protect\citeauthoryear{Kojima \bgroup et al\mbox.\egroup
  }{2022}]{kojima2022large}
Kojima, T.; Gu, S.~S.; Reid, M.; Matsuo, Y.; and Iwasawa, Y.
\newblock 2022.
\newblock Large language models are zero-shot reasoners.
\newblock {\em Advances in neural information processing systems}
  35:22199--22213.

\bibitem[\protect\citeauthoryear{K{\"o}kciyan \bgroup et al\mbox.\egroup
  }{2021}]{kokciyan2021applying}
K{\"o}kciyan, N.; Sassoon, I.; Sklar, E.; Modgil, S.; and Parsons, S.
\newblock 2021.
\newblock Applying metalevel argumentation frameworks to support medical
  decision making.
\newblock {\em IEEE Intelligent Systems} 36(2):64--71.

\bibitem[\protect\citeauthoryear{Lee, Goldwasser, and
  Reese}{2024}]{lee-etal-2024-towards}
Lee, Y.; Goldwasser, D.; and Reese, L.~S.
\newblock 2024.
\newblock Towards understanding counseling conversations: Domain knowledge and
  large language models.
\newblock In Graham, Y., and Purver, M., eds., {\em Findings of the Association
  for Computational Linguistics: EACL 2024},  2032--2047.
\newblock St. Julian{'}s, Malta: Association for Computational Linguistics.

\bibitem[\protect\citeauthoryear{Lewkowycz \bgroup et al\mbox.\egroup
  }{2022}]{lewkowycz2022solving}
Lewkowycz, A.; Andreassen, A.; Dohan, D.; Dyer, E.; Michalewski, H.; Ramasesh,
  V.; Slone, A.; Anil, C.; Schlag, I.; Gutman-Solo, T.; et~al.
\newblock 2022.
\newblock Solving quantitative reasoning problems with language models.
\newblock {\em Advances in Neural Information Processing Systems}
  35:3843--3857.

\bibitem[\protect\citeauthoryear{Liu and Chilton}{2022}]{liu2022design}
Liu, V., and Chilton, L.~B.
\newblock 2022.
\newblock Design guidelines for prompt engineering text-to-image generative
  models.
\newblock In {\em Proceedings of the 2022 CHI Conference on Human Factors in
  Computing Systems},  1--23.

\bibitem[\protect\citeauthoryear{Liu \bgroup et al\mbox.\egroup
  }{2023}]{liu2023pre}
Liu, P.; Yuan, W.; Fu, J.; Jiang, Z.; Hayashi, H.; and Neubig, G.
\newblock 2023.
\newblock Pre-train, prompt, and predict: A systematic survey of prompting
  methods in natural language processing.
\newblock {\em ACM Computing Surveys} 55(9):1--35.

\bibitem[\protect\citeauthoryear{Liu \bgroup et al\mbox.\egroup
  }{2024}]{liu2024llm}
Liu, N.; Chen, L.; Tian, X.; Zou, W.; Chen, K.; and Cui, M.
\newblock 2024.
\newblock From llm to conversational agent: A memory enhanced architecture with
  fine-tuning of large language models.
\newblock {\em arXiv preprint arXiv:2401.02777}.

\bibitem[\protect\citeauthoryear{Lu \bgroup et al\mbox.\egroup
  }{2023}]{lu2023memochat}
Lu, J.; An, S.; Lin, M.; Pergola, G.; He, Y.; Yin, D.; Sun, X.; and Wu, Y.
\newblock 2023.
\newblock Memochat: Tuning llms to use memos for consistent long-range
  open-domain conversation.
\newblock {\em arXiv preprint arXiv:2308.08239}.

\bibitem[\protect\citeauthoryear{MacAskill, Bykvist, and
  Ord}{2020}]{macaskill2020moral}
MacAskill, M.; Bykvist, K.; and Ord, T.
\newblock 2020.
\newblock {\em Moral uncertainty}.
\newblock Oxford University Press.

\bibitem[\protect\citeauthoryear{Modgil and Caminada}{2009}]{ModCam}
Modgil, S., and Caminada, M.
\newblock 2009.
\newblock Chapter 6 : Proof theories and algorithms for abstract argumentation
  frameworks.
\newblock In Simari, G., and Rahwan, I., eds., {\em Argumentation in AI}.
  Springer.
\newblock  105--129.

\bibitem[\protect\citeauthoryear{Modgil and Prakken}{2010}]{EAFandASPIC}
Modgil, S., and Prakken, H.
\newblock 2010.
\newblock Reasoning about preferences in structured extended argumentation
  frameworks.
\newblock In {\em Proceedings of the 2010 Conference on Computational Models of
  Argument: Proceedings of COMMA 2010},  347–358.
\newblock NLD: IOS Press.

\bibitem[\protect\citeauthoryear{Modgil and Prakken}{2013}]{modgil2013general}
Modgil, S., and Prakken, H.
\newblock 2013.
\newblock A general account of argumentation with preferences.
\newblock {\em Artificial Intelligence} 195:361--397.

\bibitem[\protect\citeauthoryear{Modgil and Prakken}{2018}]{modgil2018abstract}
Modgil, S., and Prakken, H.
\newblock 2018.
\newblock Abstract rule--based argumentation.
\newblock In P.~Baroni, D.~Gabbay, M.~G., and van~der Torre, L., eds., {\em
  Handbook of Formal Argumentation}.
\newblock  286--361.

\bibitem[\protect\citeauthoryear{Modgil}{2017a}]{modgil2017dialogical}
Modgil, S.
\newblock 2017a.
\newblock Dialogical scaffolding for human and artificial agent reasoning.
\newblock In {\em AIC},  58--71.

\bibitem[\protect\citeauthoryear{Modgil}{2017b}]{modgil2017towards}
Modgil, S.
\newblock 2017b.
\newblock Towards a general framework for dialogues that accommodate reasoning
  about preferences.
\newblock In {\em International Workshop on Theorie and Applications of Formal
  Argumentation},  175--191.
\newblock Springer.

\bibitem[\protect\citeauthoryear{Papachristou, Yang, and
  Hsu}{2023}]{papachristou2023leveraging}
Papachristou, M.; Yang, L.; and Hsu, C.-C.
\newblock 2023.
\newblock Leveraging large language models for collective decision-making.
\newblock {\em arXiv preprint arXiv:2311.04928}.

\bibitem[\protect\citeauthoryear{Prakken}{2005}]{prakken2005coherence}
Prakken, H.
\newblock 2005.
\newblock Coherence and flexibility in dialogue games for argumentation.
\newblock {\em Journal of logic and computation} 15(6):1009--1040.

\bibitem[\protect\citeauthoryear{Prakken}{2009}]{Prakken2009}
Prakken, H.
\newblock 2009.
\newblock {\em Models of Persuasion Dialogue}.
\newblock Boston, MA: Springer US.
\newblock  281--300.

\bibitem[\protect\citeauthoryear{Rajani \bgroup et al\mbox.\egroup
  }{2019}]{rajani2019explain}
Rajani, N.~F.; McCann, B.; Xiong, C.; and Socher, R.
\newblock 2019.
\newblock Explain yourself! leveraging language models for commonsense
  reasoning.
\newblock {\em arXiv preprint arXiv:1906.02361}.

\bibitem[\protect\citeauthoryear{Russell, Dewey, and
  Tegmark}{2015}]{russell2015research}
Russell, S.; Dewey, D.; and Tegmark, M.
\newblock 2015.
\newblock Research priorities for robust and beneficial artificial
  intelligence.
\newblock {\em AI magazine} 36(4):105--114.

\bibitem[\protect\citeauthoryear{Russell}{2020}]{HumanCompatible}
Russell, S.
\newblock 2020.
\newblock {\em Human Compatible: AI and the Problem of Control}.
\newblock Penguin Press.

\bibitem[\protect\citeauthoryear{Sumers \bgroup et al\mbox.\egroup
  }{2023}]{sumers2023cognitive}
Sumers, T.~R.; Yao, S.; Narasimhan, K.; and Griffiths, T.~L.
\newblock 2023.
\newblock Cognitive architectures for language agents.
\newblock {\em arXiv preprint arXiv:2309.02427}.

\bibitem[\protect\citeauthoryear{Suzgun \bgroup et al\mbox.\egroup
  }{2022}]{suzgun2022challenging}
Suzgun, M.; Scales, N.; Sch{\"a}rli, N.; Gehrmann, S.; Tay, Y.; Chung, H.~W.;
  Chowdhery, A.; Le, Q.~V.; Chi, E.~H.; Zhou, D.; et~al.
\newblock 2022.
\newblock Challenging big-bench tasks and whether chain-of-thought can solve
  them.
\newblock {\em arXiv preprint arXiv:2210.09261}.

\bibitem[\protect\citeauthoryear{Szabo \bgroup et al\mbox.\egroup
  }{2024}]{szabo2024moral}
Szabo, J.; Criado, N.; Such, J.; and Modgil, S.
\newblock 2024.
\newblock Moral uncertainty and the problem of fanaticism.
\newblock In {\em Proceedings of the AAAI Conference on Artificial
  Intelligence}, volume~38,  19948--19955.

\bibitem[\protect\citeauthoryear{Taylor \bgroup et al\mbox.\egroup
  }{2020}]{Alignment}
Taylor, J.; Yudkowsky, E.; LaVictoire, P.; and Critch, A.
\newblock 2020.
\newblock {\em Alignment for Advanced Machine Learning Systems}.
\newblock  342--382.

\bibitem[\protect\citeauthoryear{Thang and Luong}{2014}]{ThangPS}
Thang, P.~M., and Luong, H.~T.
\newblock 2014.
\newblock Translating preferred subtheories into structured argumentation.
\newblock {\em Journal of Logic and Computation} 74(4):831--849.

\bibitem[\protect\citeauthoryear{Tolchinsky \bgroup et al\mbox.\egroup
  }{2012}]{tolchinsky2012deliberation}
Tolchinsky, P.; Modgil, S.; Atkinson, K.; McBurney, P.; and Cort{\'e}s, U.
\newblock 2012.
\newblock Deliberation dialogues for reasoning about safety critical actions.
\newblock {\em Autonomous Agents and Multi-Agent Systems} 25(2):209--259.

\bibitem[\protect\citeauthoryear{Touvron \bgroup et al\mbox.\egroup
  }{2023}]{touvron2023llama}
Touvron, H.; Lavril, T.; Izacard, G.; Martinet, X.; Lachaux, M.-A.; Lacroix,
  T.; Rozi{\`e}re, B.; Goyal, N.; Hambro, E.; Azhar, F.; et~al.
\newblock 2023.
\newblock Llama: Open and efficient foundation language models.
\newblock {\em arXiv preprint arXiv:2302.13971}.

\bibitem[\protect\citeauthoryear{Turpin \bgroup et al\mbox.\egroup
  }{2024}]{turpin2024language}
Turpin, M.; Michael, J.; Perez, E.; and Bowman, S.
\newblock 2024.
\newblock Language models don't always say what they think: unfaithful
  explanations in chain-of-thought prompting.
\newblock {\em Advances in Neural Information Processing Systems} 36.

\bibitem[\protect\citeauthoryear{Visser \bgroup et al\mbox.\egroup
  }{2021}]{viss}
Visser, J.; Lawrence, J.; Reed, C.; Wagemans, J.; and Walton, D.
\newblock 2021.
\newblock Annotating argument schemes.
\newblock {\em Argumentation} 35(1):101--139.

\bibitem[\protect\citeauthoryear{Vlastos}{1982}]{vlastos1982socratic}
Vlastos, G.
\newblock 1982.
\newblock The socratic elenchus.
\newblock {\em The Journal of Philosophy} 79(11):711--714.

\bibitem[\protect\citeauthoryear{Wachowiak and
  Gromann}{2023}]{wachowiak2023does}
Wachowiak, L., and Gromann, D.
\newblock 2023.
\newblock Does gpt-3 grasp metaphors? identifying metaphor mappings with
  generative language models.
\newblock In {\em Proceedings of the 61st Annual Meeting of the Association for
  Computational Linguistics (Volume 1: Long Papers)},  1018--1032.

\bibitem[\protect\citeauthoryear{Walton and Macagno}{2015}]{waltmac}
Walton, D., and Macagno, F.
\newblock 2015.
\newblock A classification system for argumentation schemes.
\newblock {\em Argument \& Computation} 6(3):219--245.

\bibitem[\protect\citeauthoryear{Walton, Reed, and Macagno}{2008}]{walt}
Walton, D.; Reed, C.; and Macagno, F.
\newblock 2008.
\newblock {\em Argumentation schemes}.
\newblock Cambridge University Press.

\bibitem[\protect\citeauthoryear{Walton}{2009}]{walton2009argumentation}
Walton, D.
\newblock 2009.
\newblock Argumentation theory: A very short introduction.
\newblock In {\em Argumentation in artificial intelligence}. Springer.
\newblock  1--22.

\bibitem[\protect\citeauthoryear{Wang \bgroup et al\mbox.\egroup
  }{2022}]{wang2022self}
Wang, X.; Wei, J.; Schuurmans, D.; Le, Q.; Chi, E.; Narang, S.; Chowdhery, A.;
  and Zhou, D.
\newblock 2022.
\newblock Self-consistency improves chain of thought reasoning in language
  models.
\newblock {\em arXiv preprint arXiv:2203.11171}.

\bibitem[\protect\citeauthoryear{Wang \bgroup et al\mbox.\egroup
  }{2023}]{wang2023survey}
Wang, L.; Ma, C.; Feng, X.; Zhang, Z.; Yang, H.; Zhang, J.; Chen, Z.; Tang, J.;
  Chen, X.; Lin, Y.; et~al.
\newblock 2023.
\newblock A survey on large language model based autonomous agents.
\newblock {\em arXiv preprint arXiv:2308.11432}.

\bibitem[\protect\citeauthoryear{Wei \bgroup et al\mbox.\egroup
  }{2022a}]{wei2022emergent}
Wei, J.; Tay, Y.; Bommasani, R.; Raffel, C.; Zoph, B.; Borgeaud, S.; Yogatama,
  D.; Bosma, M.; Zhou, D.; Metzler, D.; et~al.
\newblock 2022a.
\newblock Emergent abilities of large language models.
\newblock {\em arXiv preprint arXiv:2206.07682}.

\bibitem[\protect\citeauthoryear{Wei \bgroup et al\mbox.\egroup
  }{2022b}]{wei2022chain}
Wei, J.; Wang, X.; Schuurmans, D.; Bosma, M.; Xia, F.; Chi, E.; Le, Q.~V.;
  Zhou, D.; et~al.
\newblock 2022b.
\newblock Chain-of-thought prompting elicits reasoning in large language
  models.
\newblock {\em Advances in neural information processing systems}
  35:24824--24837.

\bibitem[\protect\citeauthoryear{White \bgroup et al\mbox.\egroup
  }{2023}]{white2023prompt}
White, J.; Fu, Q.; Hays, S.; Sandborn, M.; Olea, C.; Gilbert, H.; Elnashar, A.;
  Spencer-Smith, J.; and Schmidt, D.~C.
\newblock 2023.
\newblock A prompt pattern catalog to enhance prompt engineering with chatgpt.
\newblock {\em arXiv preprint arXiv:2302.11382}.

\bibitem[\protect\citeauthoryear{Wu \bgroup et al\mbox.\egroup
  }{2023}]{wu2023bloomberggpt}
Wu, S.; Irsoy, O.; Lu, S.; Dabravolski, V.; Dredze, M.; Gehrmann, S.; Kambadur,
  P.; Rosenberg, D.; and Mann, G.
\newblock 2023.
\newblock Bloomberggpt: A large language model for finance.
\newblock {\em arXiv preprint arXiv:2303.17564}.

\bibitem[\protect\citeauthoryear{Xi \bgroup et al\mbox.\egroup
  }{2023}]{xi2023rise}
Xi, Z.; Chen, W.; Guo, X.; He, W.; Ding, Y.; Hong, B.; Zhang, M.; Wang, J.;
  Jin, S.; Zhou, E.; et~al.
\newblock 2023.
\newblock The rise and potential of large language model based agents: A
  survey.
\newblock {\em arXiv preprint arXiv:2309.07864}.

\bibitem[\protect\citeauthoryear{Xydis \bgroup et al\mbox.\egroup
  }{2020}]{xydis2020enthymemes}
Xydis, A.; Hampson, C.; Modgil, S.; and Black, E.
\newblock 2020.
\newblock Enthymemes in dialogues.
\newblock In {\em COMMA},  355--369.

\bibitem[\protect\citeauthoryear{Xydis \bgroup et al\mbox.\egroup
  }{2022}]{xydis2022sound}
Xydis, A.; Hampson, C.; Modgil, S.; and Black, E.
\newblock 2022.
\newblock A sound and complete dialogue system for handling misunderstandings.
\newblock In {\em SAFA@COMMA},  19--32.

\bibitem[\protect\citeauthoryear{Yang \bgroup et al\mbox.\egroup
  }{2023}]{yang2023foundation}
Yang, S.; Nachum, O.; Du, Y.; Wei, J.; Abbeel, P.; and Schuurmans, D.
\newblock 2023.
\newblock Foundation models for decision making: Problems, methods, and
  opportunities.
\newblock {\em arXiv preprint arXiv:2303.04129}.

\bibitem[\protect\citeauthoryear{Young, Modgil, and Rodrigues}{2016}]{young16}
Young, A.; Modgil, S.; and Rodrigues, O.
\newblock 2016.
\newblock Prioritised default logic as rational argumentation.
\newblock In {\em Proc. 15th Int. Conference on Autonomous Agents and
  Multiagent Systems (AAMAS16)},  626--634.

\bibitem[\protect\citeauthoryear{Zelikman \bgroup et al\mbox.\egroup
  }{2022}]{zelikman2022star}
Zelikman, E.; Wu, Y.; Mu, J.; and Goodman, N.
\newblock 2022.
\newblock {ST}ar: Bootstrapping reasoning with reasoning.
\newblock In Oh, A.~H.; Agarwal, A.; Belgrave, D.; and Cho, K., eds., {\em
  Advances in Neural Information Processing Systems}.

\bibitem[\protect\citeauthoryear{Zhang \bgroup et al\mbox.\egroup
  }{2022a}]{zhang2022opt}
Zhang, S.; Roller, S.; Goyal, N.; Artetxe, M.; Chen, M.; Chen, S.; Dewan, C.;
  Diab, M.; Li, X.; Lin, X.~V.; et~al.
\newblock 2022a.
\newblock Opt: Open pre-trained transformer language models.
\newblock {\em arXiv preprint arXiv:2205.01068}.

\bibitem[\protect\citeauthoryear{Zhang \bgroup et al\mbox.\egroup
  }{2022b}]{zhang2022automatic}
Zhang, Z.; Zhang, A.; Li, M.; and Smola, A.
\newblock 2022b.
\newblock Automatic chain of thought prompting in large language models.
\newblock {\em arXiv preprint arXiv:2210.03493}.

\bibitem[\protect\citeauthoryear{Zhang \bgroup et al\mbox.\egroup
  }{2024}]{zhang2024survey}
Zhang, Z.; Bo, X.; Ma, C.; Li, R.; Chen, X.; Dai, Q.; Zhu, J.; Dong, Z.; and
  Wen, J.-R.
\newblock 2024.
\newblock A survey on the memory mechanism of large language model based
  agents.
\newblock {\em arXiv preprint arXiv:2404.13501}.

\bibitem[\protect\citeauthoryear{Zhao \bgroup et al\mbox.\egroup
  }{2023}]{zhao2023survey}
Zhao, W.~X.; Zhou, K.; Li, J.; Tang, T.; Wang, X.; Hou, Y.; Min, Y.; Zhang, B.;
  Zhang, J.; Dong, Z.; et~al.
\newblock 2023.
\newblock A survey of large language models.
\newblock {\em arXiv preprint arXiv:2303.18223}.

\end{thebibliography}

\end{document}